\documentclass{svproc}

\usepackage{graphicx}
\usepackage{subfig}
\usepackage{amsmath}
\usepackage{multicol}

\newcommand{\squeezeupbig}{\vspace{-5mm}}
\newcommand{\squeezeupsmall}{\vspace{-2.5mm}}

\DeclareMathOperator*{\argmin}{arg\,min}

\usepackage{url}

\begin{document}

\title{Style transfer with adaptation to the central objects of the scene}
\author{Alexey A. Schekalev\inst{1}, Victor V. Kitov\inst{1,2}}
\institute{Lomonosov Moscow State University, e-mail \email{Alexey.Schekalev@sas.com},
\and
Plekhanov Russian University of Economics, e-mail: \email{v.v.kitov@yandex.ru}
}
\maketitle
\begin{abstract}
Style transfer is a problem of rendering  image with some content in the style of another image, for example a family photo in the style of a painting of some famous artist. The drawback of classical style transfer algorithm is that it imposes style uniformly on all parts of the content image, which perturbs central objects on the content image, such as faces or text, and makes them unrecognizable.
This work proposes a novel style transfer algorithm which automatically detects central objects on the content image, generates spatial importance mask and imposes style non-uniformly: central objects are stylized less to preserve their recognizability and other parts of the image are stylized as usual to preserve the style.
Three methods of automatic central object detection are proposed and evaluated qualitatively and via a user evaluation study. Both comparisons demonstrate higher quality of stylization compared to the classical style transfer method.

\end{abstract}

\section{Introduction}
\begin{figure}[ht]
	\centering
	\includegraphics[width=0.8\textwidth]{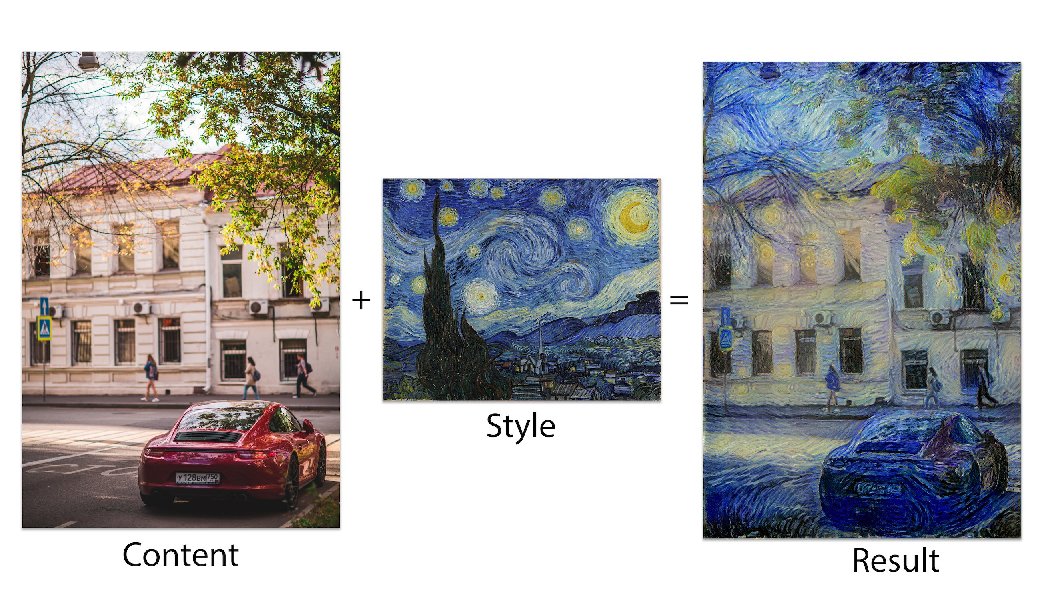}
	\caption{Style transer task}
	\label{common_task}
	\squeezeupbig
\end{figure}

Image stylization~\cite{adobe} is a classical problem in computer vision of rendering a content image in the style of another style image, as shown on Fig. \ref{common_task}. Earlier approaches used hard-coded rules to impose predefined style. Recently, a method of Gatys et al.\cite{Gatys} was proposed to impose arbitrary style on arbitrary content image using deep convolutional networks. 

The main task is to transfer style from one image to another. This algorithm should work with any content and style images. In 2016 Leon Gatys proposed a method \cite{Gatys} of stylization, based on deep neural networks, which solved this problem. The main idea was to optimize in the space of images to find a picture semantically reflecting content from the content image and the style of the style image. These two contradicting goals were regulated by minimizing simultaneously content loss and style loss: 
\begin{equation}
x = \argmin\limits_{x}\{\alpha\mathcal{L}_{\text{content}}\left(x, x_c\right) + \mathcal{L}_{\text{style}}\left(x, x_s\right)\}
\label{optim}
\end{equation}

\begin{figure}%
\squeezeupbig
    \centering
    \subfloat[Style transfer for different $\alpha$]{\includegraphics[width=0.45\textwidth]{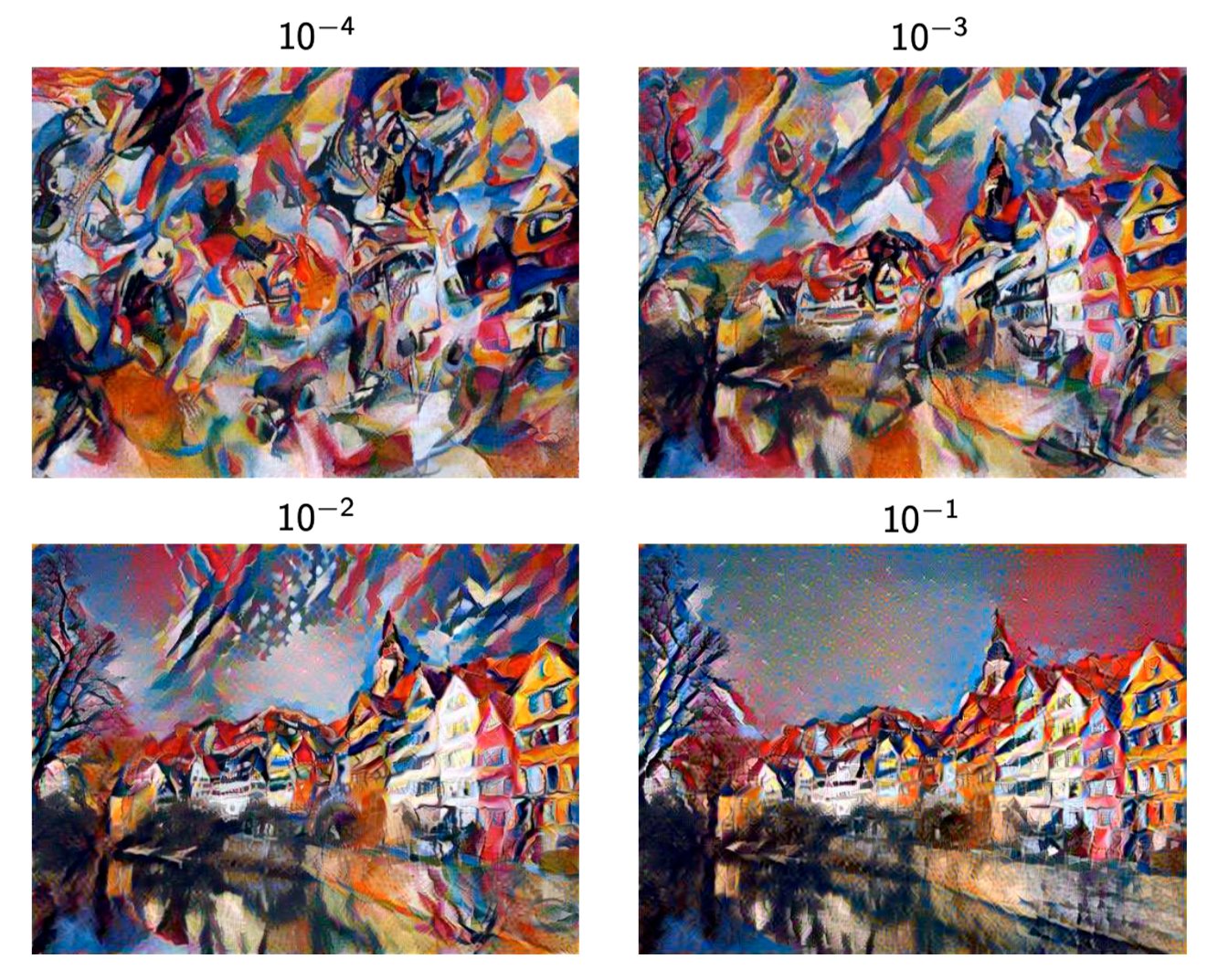}}%
    \subfloat[Problem case]{\includegraphics[width=0.55\textwidth]{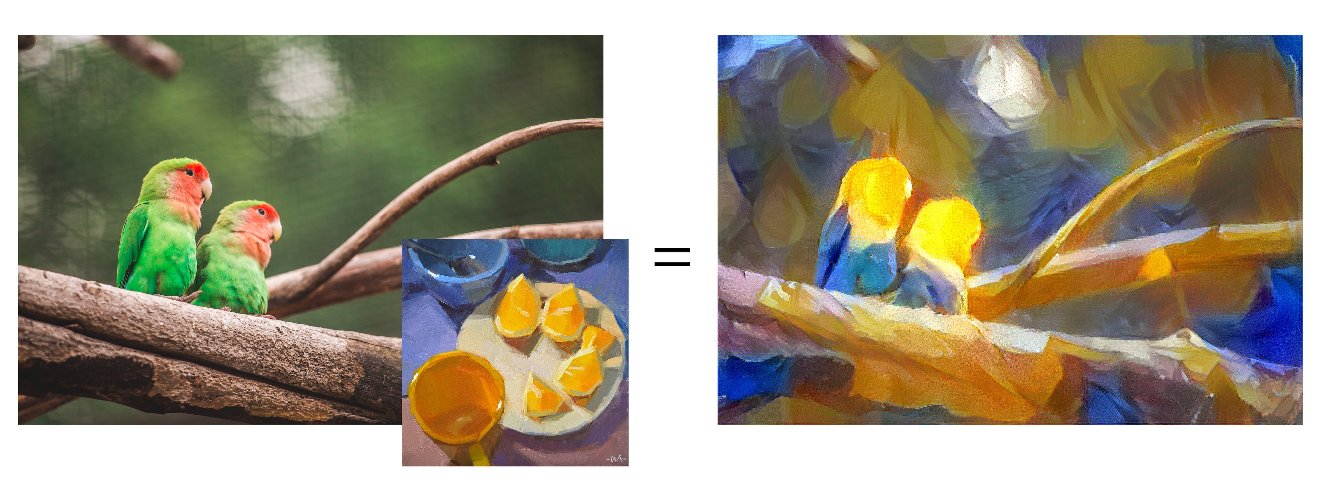} }%
    \caption{}%
    \label{ratio}%
\end{figure}

Coefficient $\alpha$ determines the strength of stylization (Fig \ref{ratio}.a). Lower $\alpha$ imposes more style and vice versa. The shortcoming of this approach is that style is imposed uniformly onto the whole content image, distorting important central objects of the image, which are critical for perception. For example, it's hard to say what bird sits on the tree (Fig. \ref{ratio}b), because small details of birds silhouette were lost during stylization.

One may improve preservation of content by increasing $\alpha$ coefficient in (\ref{optim}). However this solution decreases stylization strength globally, thus giving less expressive stylization. 

The paper proposes a new solution to this problem. First, central objects are detected and selected using automatically generated spatial importance mask for the content image. Next, this mask is used to impose style with spatially varying strength, controlled by the importance mask. This allows to achieve two contradicting goals - stylization is gentle on the central objects of the image, critical for perception, such as human faces, houses, cars, etc. And stylization is strong in the rest of the image, thus expressing a vivid style.

The paper is organized as follows. Section \ref{method} gives a description of the proposed method and provides qualitative comparisons with the baseline stylization method of Gatys et al. Section  \ref{Evaluation} provides the details of the user evaluation study and summarizes its results, highlighting the superiority of the proposed solution. Section \ref{conclusion} concludes.

\section{Method}\label{method}

\subsection{Non-uniform Stylization}
Consider the loss function in the optimization problem (\ref{optim}). In the original paper \cite{Gatys} content loss is formalized as follows:
\begin{equation}
\mathcal{L}_{\text{content}}\left(x, x_c\right) = \alpha\sum\limits_{{i,j}}\left(F^{{l}}_{{i,j}} - P^{l}_{{i,j}}\right)^2	
\end{equation}
where $F^{l}$ and $P^{l}$ are internal representations in pre-trained convolutional neural network \cite{Cnn}, which is selected to be VGG \cite{VGG}. Instead of using constant $\alpha$, we propose to use a matrix with different $\alpha_{i, j}$ values for each spatial location $(i,j)$:

\begin{equation}
\mathcal{L'}_{\text{content}}\left(x, x_c\right) = \sum\limits_{{i,j}}\alpha_{i,j}\left(F^{{l}}_{{i,j}} - P^{l}_{{i,j}}\right)^2	
\label{nonuniform}
\end{equation}

Making variable $\alpha$ allows to impose less style on central objects of the scene, critical for perception, and more style in all other areas of the image..

\subsection{Automatic Central Objects Detection}

Consider convolutional neural network pre-trained for image classification. We use VGG \cite{VGG}. Such model takes input image and outputs probability distribution for each class from the ImageNet set. We detect central objects by filling different parts of the input image with uniform color and measuring change in the output class probabilities. If key object of the image was filled, one would observe drastic change in class probabilities. On the contrary, if background was changes, class probabilities would change only slightly. Overall, the magnitude of change of class probabilities determines the importance of the filled region. This approach was used to visualize convolutional neural networks in classification problems~\cite{Visualization}, but in the problem of style transfer, to our knowledge, it is used for the first time. After splitting whole image into a set of regions and filling each region one by one and evaluating its importance, we build a whole importance map $\alpha_{i,j}$ measuring semantic significance of each location of the image. This importance map is passed to the spatially varying style transfer algorithm (\ref{optim}) with modified content loss function (\ref{nonuniform}).

\begin{figure}%
    \centering
    \subfloat[the probability distribution for input]{\includegraphics[width=0.5\textwidth]{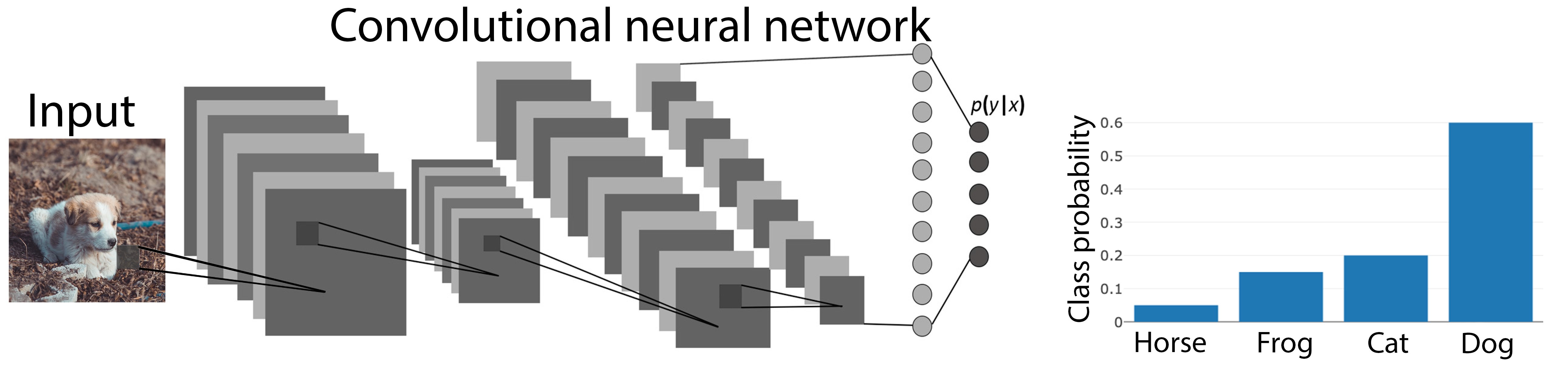}}%
    \subfloat[changing the probability distribution when the patch is overwritten]{\includegraphics[width=0.5\textwidth]{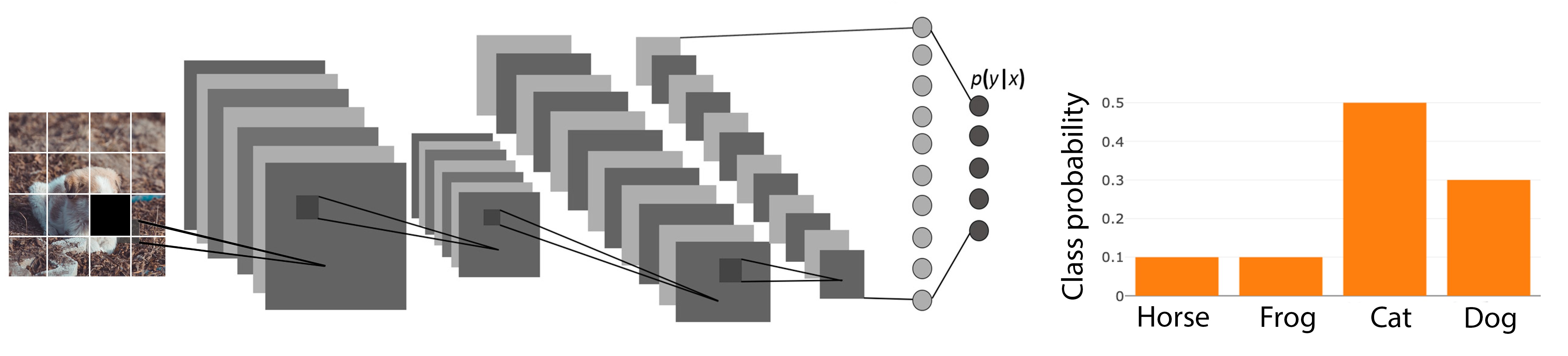} }%
    \caption{}%
    \label{cheme}%
\end{figure}

\subsubsection{Patch-Based Mask Generation} 
\begin{figure}%
	\squeezeupsmall 
    \centering 
    \subfloat[patch importance]{\includegraphics[height = 4cm]{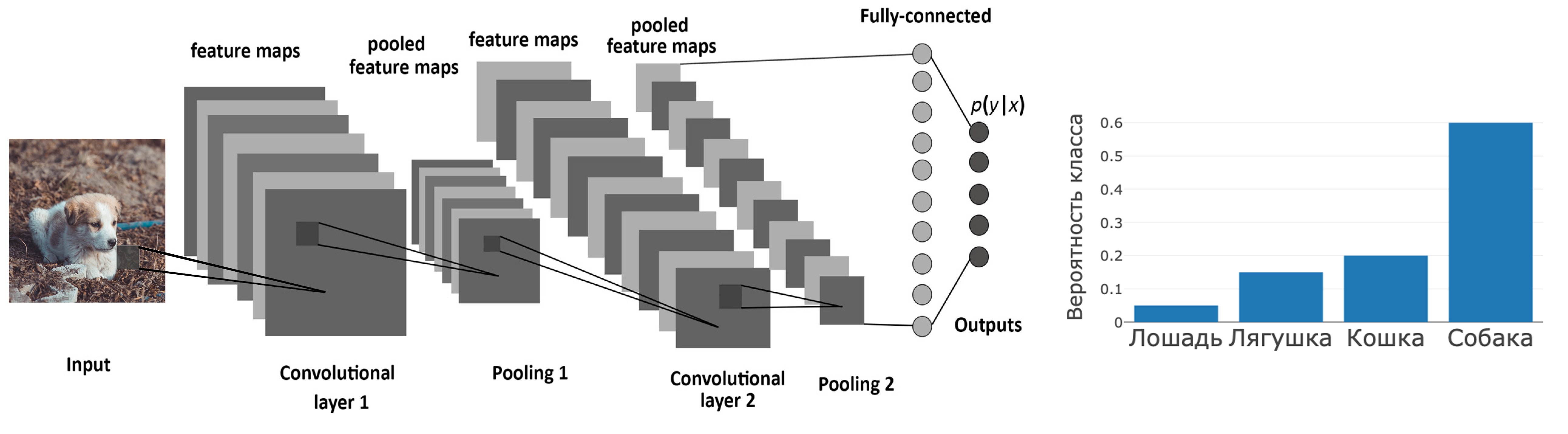}}%
    \qquad
    \subfloat[baseline]{\includegraphics[height = 4cm]{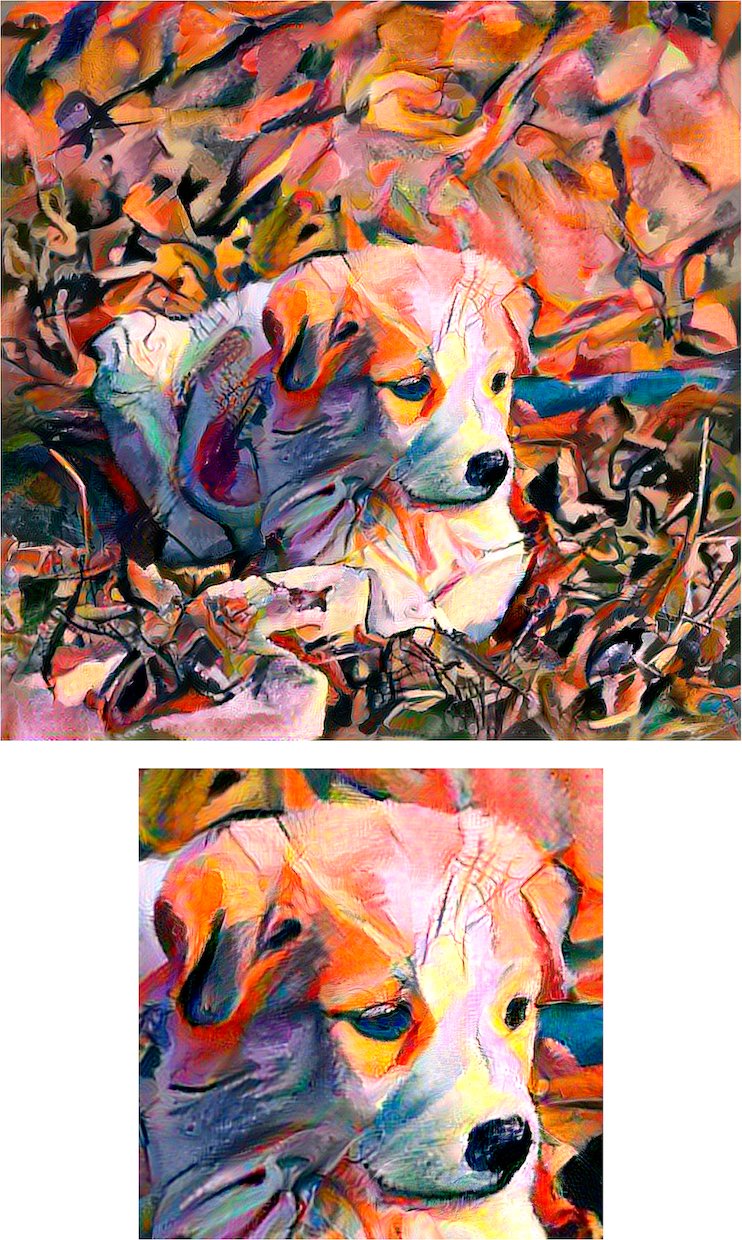}}%
    \qquad
    \subfloat[patch stylisation]{\includegraphics[height = 4cm]{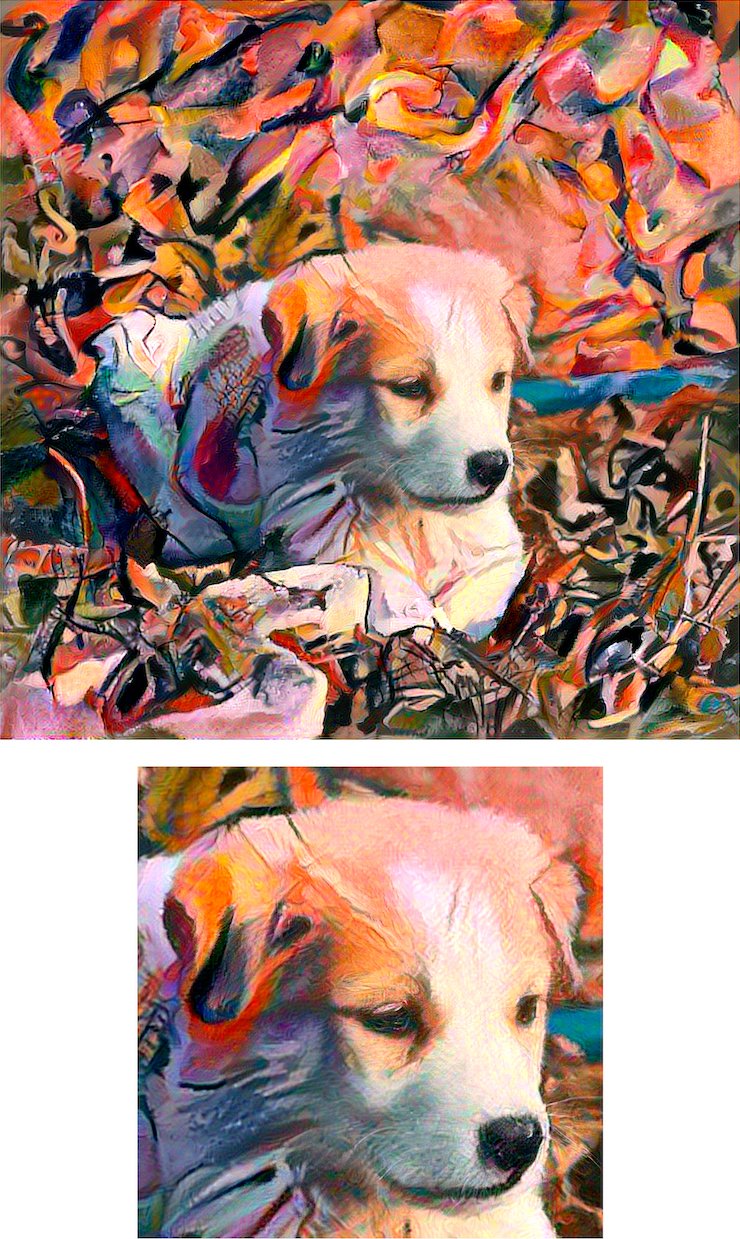} }%
    \caption{}%
    \label{importance}%
    \squeezeupbig
\end{figure}

In this approach we propose to divide the image by a uniform patch grid (like at Fig \ref{cheme}.b). Sequentially overwriting the patches and passing the image through the neural network, we rate the importance of the patches by calculating $L_2$ norm of class distributions difference. Visualization of results shows that proposed algorithm could find central object of the scene and separate them from background (Fig. \ref{importance}.a). After that we use found $\alpha_{i, j}$ matrix in stylization algorithm with changed content loss (\ref{nonuniform}). At Fig.\ref{importance} (b and c) we could see the difference between baseline approach and proposed model. There are a lot of small details at dog face failed to save in baseline approach and could save in new model. 

\begin{figure}[h]%
	\squeezeupbig
    \centering
    \subfloat[averaging $\alpha$ matrices]{\includegraphics[height = 4cm]{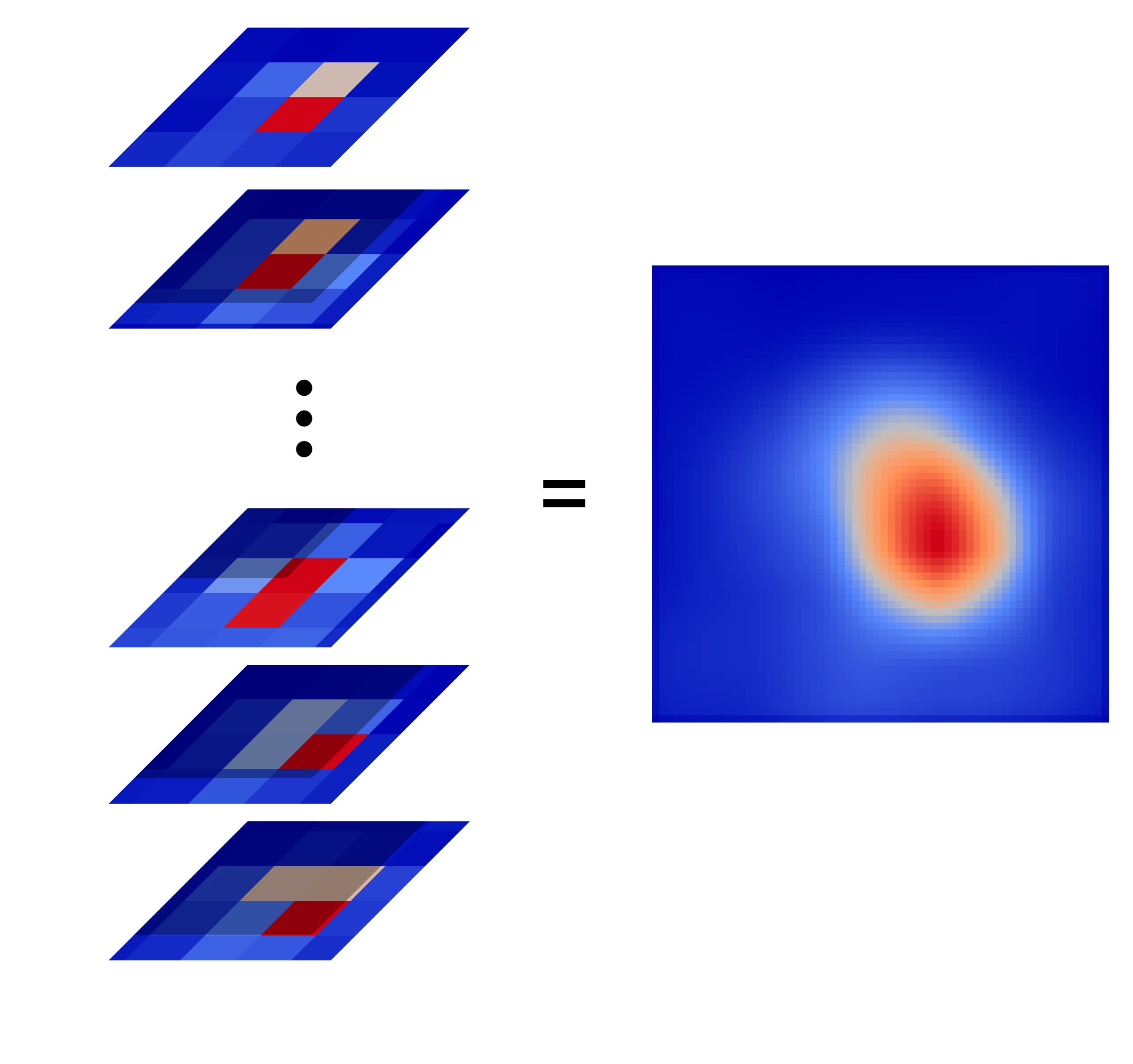}}%
    \qquad
    \subfloat[baseline]{\includegraphics[height = 4cm]{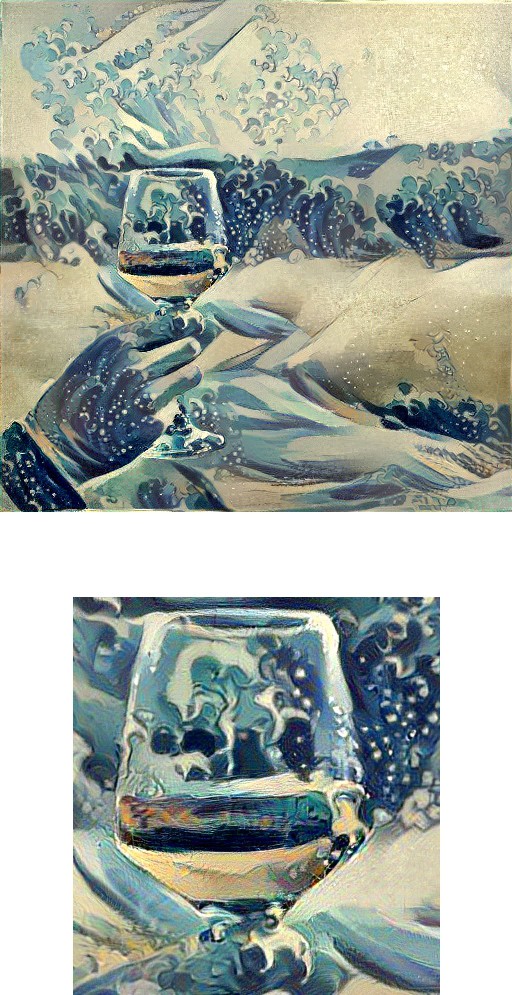}}%
    \qquad
    \subfloat[averaging patch stylization]{\includegraphics[height = 4cm]{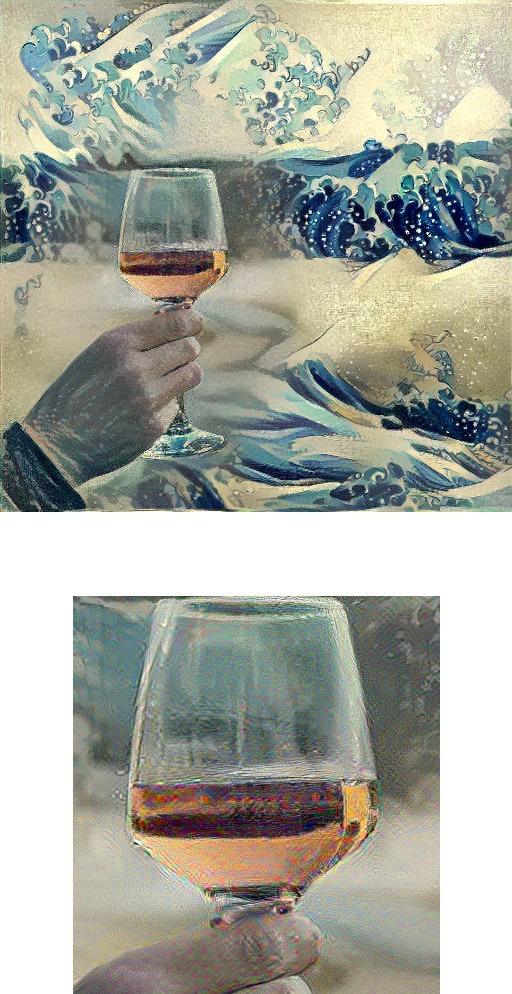}}%
    \caption{}%
    \label{avg_patch}%
    \squeezeupbig
\end{figure}

At the example above (Fig. \ref{importance}.a) we see, that main patch covers not only the central object, but it covers background too. Instead of using fixed patches we additionally propose to use previous algorithm for different position of the grid mesh and combine results together (Fig.\ref{avg_patch}a) by pixelwise averaging. We see the difference between two approaches at Fig.\ref{avg_patch}(b and c). Averaging of different matrices allows to obtain more smooth distribution of weights, so it allows to define the boundaries of central objects better.

\subsubsection{Superpixel-Based Mask Generation}
\begin{figure}%
	\squeezeupbig
    \centering
    \subfloat[superpixels]{\includegraphics[height = 2.5cm]{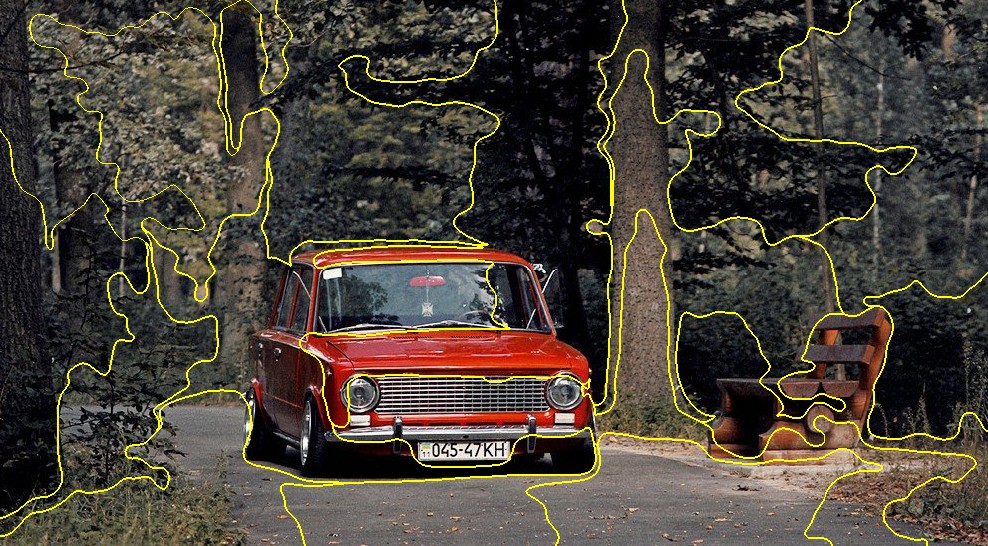}}%
    \qquad
    \subfloat[averaging $\alpha$ matrices]{\includegraphics[height = 2.5cm]{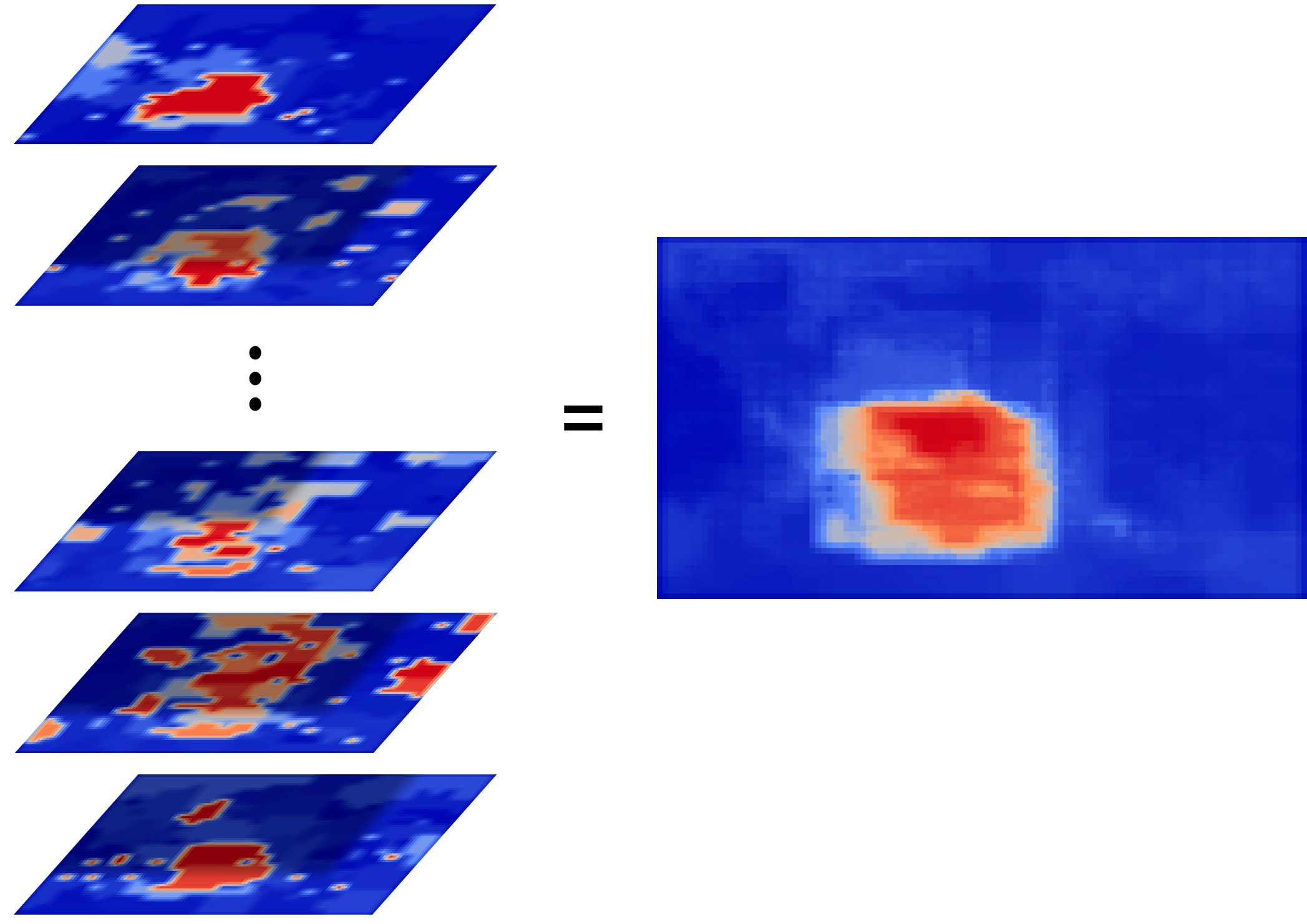}}%
    \caption{}%
    \label{superpixel}%
    \squeezeupbig
\end{figure}

At the example above (Fig. \ref{avg_patch}) we see that averaging of different $\alpha_{i,j}$ matrices produces boundary of elliptical form. If central objects have more complicated boundaries, the proposed method becomes unsuitable. To improve the results, instead of using a uniform grid, we suggest to split the image into superpixels \cite{superpixel}. This algorithm divides the image into small segments (superpixels), the boundaries of which are close to the boundaries of the objects in the image (Fig. \ref{superpixel}a). Superpixel algorithm has two main parameters, responsible for the number of segments and the shape of boundaries. We choose a set of predefined values of these parameters and run importance mask evaluation algorithm several times, then average the results for better quality (Fig. \ref{superpixel}b)

\begin{figure}%
\squeezeupbig
    \centering
    \subfloat[baseline]{\includegraphics[width=0.25\textwidth]{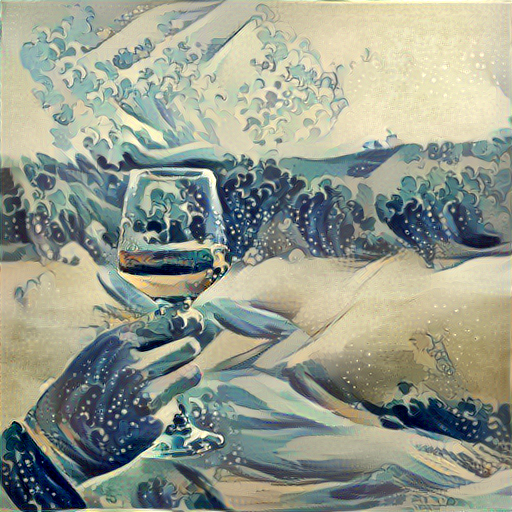}}%
    \qquad
    \subfloat[averaging patch stylization]{\includegraphics[width=0.25\textwidth]{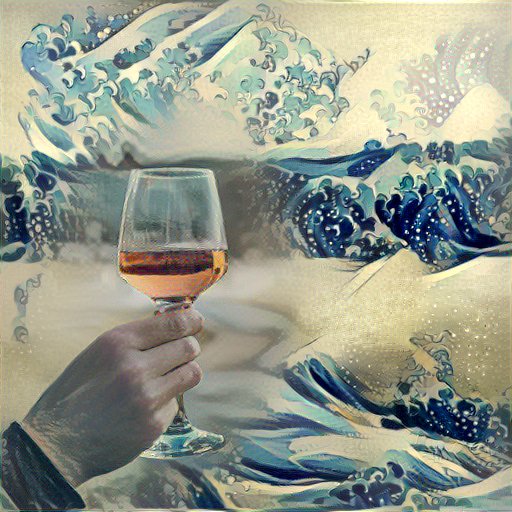}}%
    \qquad
    \subfloat[averaging superpixel stylization]{\includegraphics[width=0.25\textwidth]{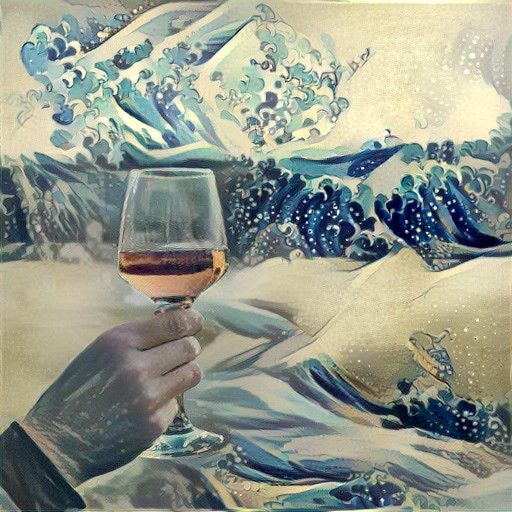}}%
    \caption{}%
    \label{comparing}%
    \squeezeupbig
\end{figure}

Fig \ref{comparing} shows qualitative difference between uniform stylization (a) and patch-based (b) and superpixel-based (c) spatially varying stylization. Boundaries of the central object -- the glass -- are non-convex, thus superpixel-based extracts the boundary of such object better, which improves the quality of final stylization. 
\subsubsection{Segmentation-Based Mask Generation}
\begin{figure}%
    \centering
    \subfloat[superpixel stylization]{\includegraphics[width=0.4\textwidth]{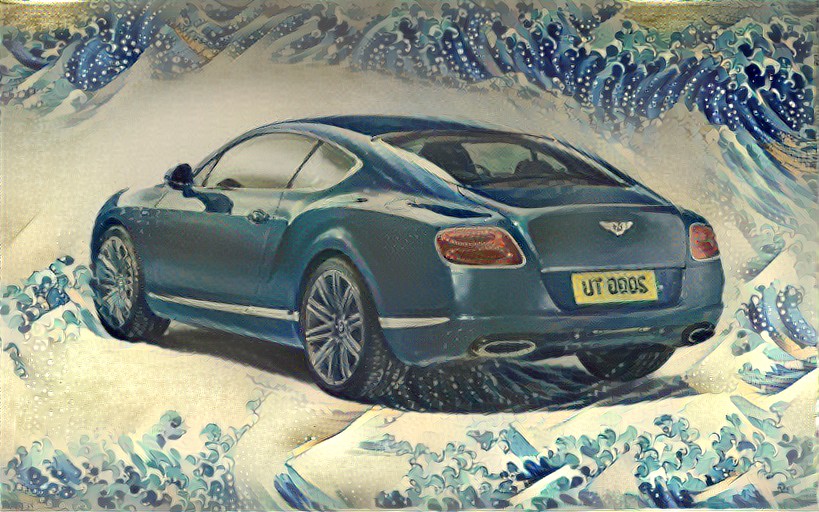}}%
    \qquad
    \subfloat[segmentation stylization]{\includegraphics[width=0.4\textwidth]{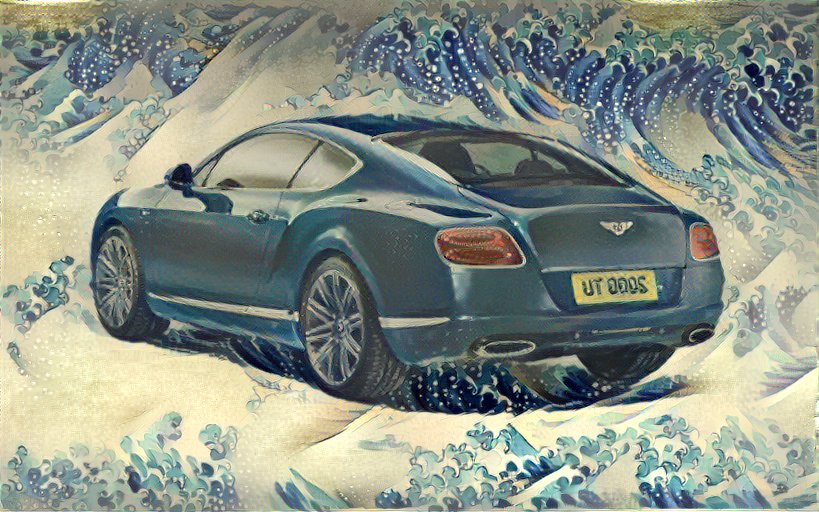}}%
    \caption{}%
    \label{bently}%
\end{figure}

Deep learning models are good at image segmentation tasks \cite{Segmentation}. So we could evaluate $\alpha_{i, j}$ matrix by previous approaches and then correct boundaries by the results of the segmentation algorithm. This approach allow to increase quality of stylization when it's easy to separate object from background. Example at fig. \ref{bently} shows, that stylization algorithm with segmentation locates the car exactly along its border, while superpixel algorithm affect some pixels near the car, which makes final style transfer less sharp along the border of the central object of the image.

\section{Results} \label{Evaluation}
To evaluate quantitatively the proposed method we a user evaluation study. For a representative set of content and style images two stylizations were obtained --- by the baseline method of Gatys et al. and by the proposed method. Respondents were asked to select a stylization they like more. To omit location bias for each comparison baseline stylization and stylization with the proposed method were shown in random order. 6 respondents were surveyed on 29 stylization outputs. We conducted 3 surveys, comparing baseline stylization algorithm, of Gatys et al. with our method, where importance mask was generated using patches, superpixels and results of image segmentation. Results are shown on table \ref{tab:surveys}.
 
 \begin{table}[]
\centering
\begin{tabular}{|l|c|}
\hline
             & Percent of vote \\ \hline
Patches      & 66            \\ \hline
Superpixel   & 72            \\ \hline
Segmentation & 80            \\ \hline
\end{tabular}\vskip1em
\caption{Comparing baseline algorithm and proposed models.}
\label{tab:surveys}
\end{table}

From these results we see that our method outperforms baseline stylization in all cases. Image segmentation modification gives maximum benefit, which can be attributed to the fact that it extracts the boundaries of central objects more accurately.

\section{Conclusion}\label{conclusion}
A new style transfer method with spatially varying strength was proposed in this work. Stylization strength was controlled for each pixel by automatically generated importance mask. Three methods, namely patch-based, segmentation-based and superpixel-based were proposed to generate importance mask. Qualitative comparisons and conducted user evaluation study demonstrated superiority of the proposed method compared to classical style transfer method of Gatys et al. due to expressive style transfer for the background and more gentle style transfer for the central objects of the content image. Among three proposed importance mask generation methods, segmentation-based showed the highest quality which may be attributed to more accurate boundary estimation of the central objects of the image.

\end{document}